\documentclass{article} 
\usepackage{iclr2027_conference,times}

\usepackage{amsmath,amsfonts,bm}

\def\eqref#1{equation~\ref{#1}}

\def\1{\bm{1}}

\DeclareMathAlphabet{\mathsfit}{\encodingdefault}{\sfdefault}{m}{sl}
\SetMathAlphabet{\mathsfit}{bold}{\encodingdefault}{\sfdefault}{bx}{n}

\usepackage{xcolor}
\usepackage{marvosym}
\definecolor{linkc}{rgb}{0,0.44,0.74}
\definecolor{eqc}{rgb}{1,0,0}
\usepackage[pagebackref=false,breaklinks=true,colorlinks=true,urlcolor=linkc,citecolor=linkc,linkcolor=eqc,bookmarks=false]{hyperref}
\usepackage{url}
\usepackage{booktabs}
\usepackage{graphicx}
\usepackage{afterpage}
\usepackage{wrapfig}
\usepackage{caption}
\title{FocusDrive: Reasoning with Visual Focus\\ for Autonomous Driving}

\author{\begin{tabular}[t]{@{}lllll@{}}
Zhiyuan Liu & Zehong Ke & Yuanxin Tian & Hao Cheng & Jinhao Li\\
Yining Xing & Yanbo Jiang & Zhenhua Xu\textsuperscript{\Letter} & Wenhao Yu\textsuperscript{\Letter} & Jianqiang Wang
\end{tabular}\\[5pt]
\normalfont Tsinghua University\\[4pt]
\normalfont\small\href{mailto:liuzhiyu24@mails.tsinghua.edu.cn}{\nolinkurl{liuzhiyu24@mails.tsinghua.edu.cn}}\\
\normalfont\small\textsuperscript{\Letter}\href{mailto:zxubg@connect.ust.hk}{\nolinkurl{zxubg@connect.ust.hk}}\quad
\textsuperscript{\Letter}\href{mailto:eyre530056@gmail.com}{\nolinkurl{eyre530056@gmail.com}}
}
\iclrfinalcopy

\newcommand{\best}[1]{\textbf{#1}}

\begin{document}
\maketitle
\lhead{Preprint}
\afterpage{\begin{figure}[t]
\centering
\includegraphics[width=\textwidth]{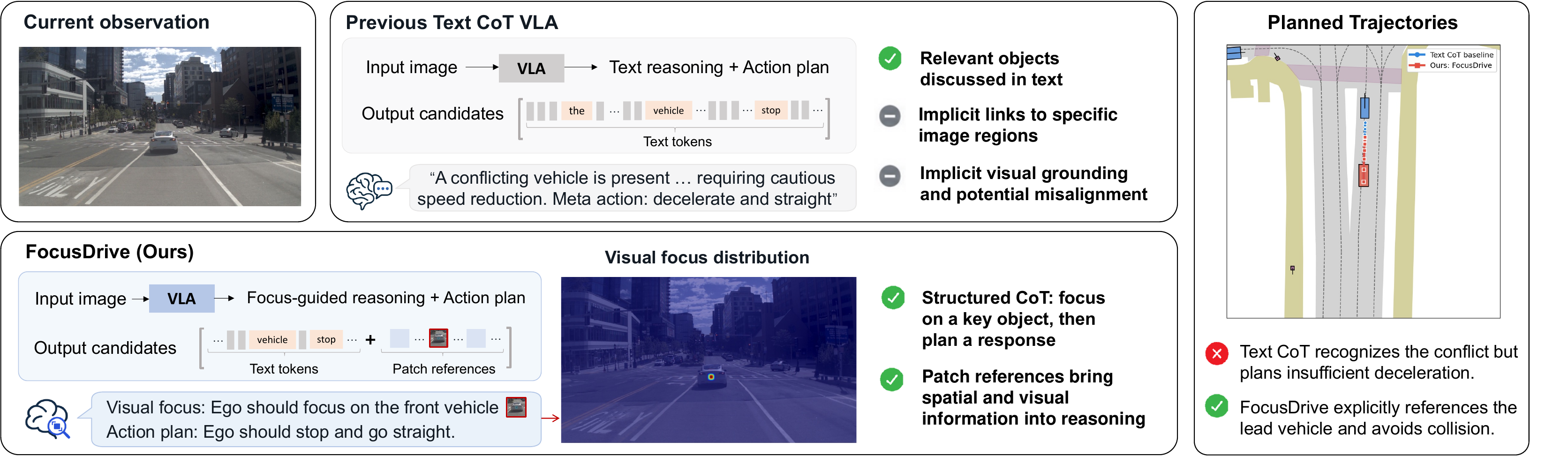}
\caption{Compared with text-only CoT, FocusDrive explicitly links objects to image patches and structures reasoning around identifying and locating a key object, then determining the appropriate driving response.}
\label{fig:teaser}
\end{figure}
}
\afterpage{\afterpage{\afterpage{\begin{figure}[t]
\centering
\includegraphics[width=\textwidth]{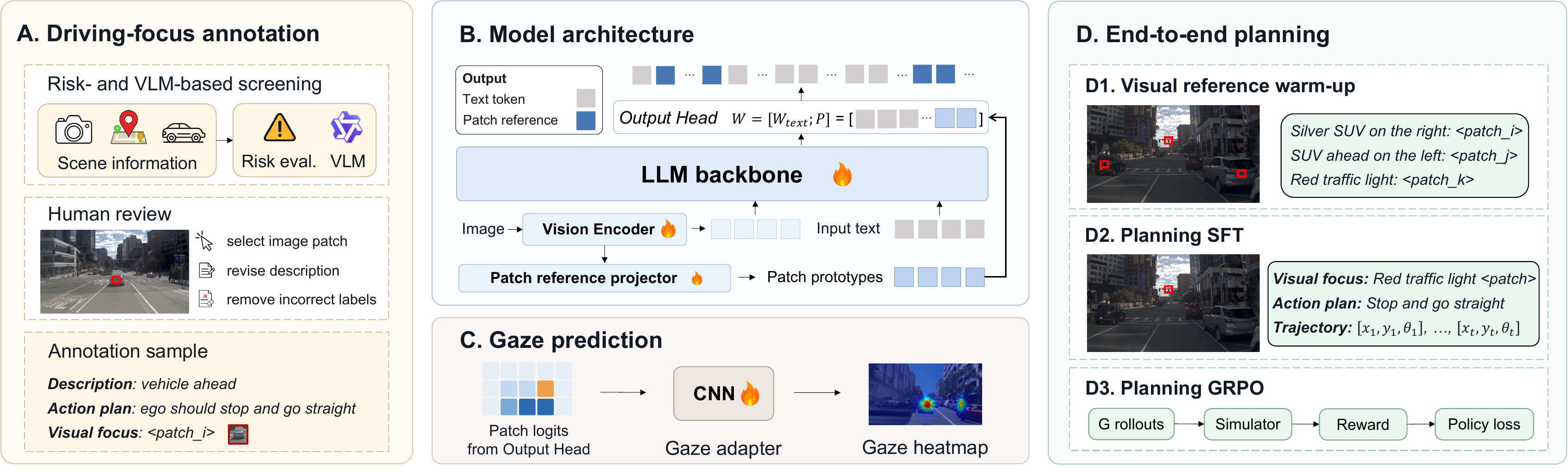}
\caption{FocusDrive overview. Risk- and VLM-based screening followed by human review provides driving-focus annotations. Image-patch references support gaze prediction through an adapter and structured focus-to-action reasoning for planning, trained with reference warm-up, SFT, and PDMS-based GRPO.}
\label{fig:pipeline}
\end{figure}
}}}
\afterpage{\afterpage{\afterpage{\afterpage{\afterpage{\afterpage{\begin{table*}[t]
\caption{Driver attention prediction across Normal, Safety, and Accident scenarios. Bold and underlined values indicate the best and second-best results. $\dagger$ denotes joint attention and language modeling.}
\label{tab:gaze-main}
\centering\scriptsize
\setlength{\tabcolsep}{2.5pt}
\renewcommand{\arraystretch}{1.0}
\resizebox{0.96\textwidth}{!}{%
\begin{tabular}{l*{15}{r}}
\toprule
& \multicolumn{5}{c}{Normal} & \multicolumn{5}{c}{Safety} & \multicolumn{5}{c}{Accident} \\
\cmidrule(lr){2-6}\cmidrule(lr){7-11}\cmidrule(lr){12-16}
Method & KLdiv$\downarrow$ & CC$\uparrow$ & SIM$\uparrow$ & AUC-J$\uparrow$ & AUC-B$\uparrow$ & KLdiv$\downarrow$ & CC$\uparrow$ & SIM$\uparrow$ & AUC-J$\uparrow$ & AUC-B$\uparrow$ & KLdiv$\downarrow$ & CC$\uparrow$ & SIM$\uparrow$ & AUC-J$\uparrow$ & AUC-B$\uparrow$ \\
\midrule
\multicolumn{16}{l}{\textit{Conventional attention predictors}} \\
MLNet~\citep{MLNet} & 2.129 & 0.547 & 0.460 & 0.914 & 0.836 & 1.953 & 0.528 & 0.433 & 0.928 & 0.874 & 2.897 & 0.344 & 0.288 & 0.893 & 0.784 \\
ConvNeXt~\citep{ConvNext} & 2.042 & 0.570 & 0.412 & 0.916 & 0.848 & 1.765 & 0.567 & 0.413 & 0.938 & 0.877 & 3.049 & 0.377 & 0.248 & 0.891 & 0.806 \\
ERFNet~\citep{ERFNet} & 1.979 & 0.558 & 0.425 & 0.923 & 0.840 & 1.593 & 0.538 & 0.410 & 0.942 & 0.868 & 2.181 & 0.391 & 0.253 & 0.930 & 0.846 \\
\midrule
\multicolumn{16}{l}{\textit{Vision--language methods}} \\
GazeXplain$\dagger$~\citep{xianyu2024gazexplain} & 2.578 & 0.477 & 0.389 & 0.857 & 0.866 & 2.769 & 0.383 & 0.321 & 0.848 & 0.743 & 3.109 & 0.371 & 0.236 & 0.902 & 0.804 \\
LLada$\dagger$~\citep{zhou2025where} & \underline{1.219} & 0.583 & 0.436 & 0.952 & \underline{0.908} & 1.230 & 0.579 & 0.420 & 0.950 & 0.912 & 1.927 & 0.396 & 0.262 & 0.934 & 0.889 \\
FSDAM$\dagger$~\citep{hamid2025fsdam} & 1.408 & 0.577 & 0.467 & 0.959 & 0.862 & 1.222 & 0.589 & 0.467 & 0.958 & 0.890 & 1.838 & 0.439 & 0.323 & 0.949 & 0.872 \\
DualGaze-VLM$\dagger$~\citep{ke2026dualgaze} & 1.229 & \underline{0.613} & \underline{0.468} & \underline{0.964} & 0.900 & \underline{1.106} & \underline{0.618} & \best{0.550} & \underline{0.961} & \underline{0.918} & \underline{1.700} & \underline{0.471} & \underline{0.325} & \underline{0.952} & \underline{0.898} \\
\textbf{FocusDrive (Ours)}$\dagger$ & \best{1.137} & \best{0.639} & \best{0.491} & \best{0.969} & \best{0.917} & \best{0.959} & \best{0.670} & \underline{0.511} & \best{0.971} & \best{0.937} & \best{1.561} & \best{0.515} & \best{0.357} & \best{0.961} & \best{0.922} \\
\bottomrule
\end{tabular}}
\end{table*}

}}}}}}
\afterpage{\afterpage{\afterpage{\afterpage{\afterpage{\afterpage{\afterpage{\begin{table*}[t]
\caption{Planning results on NAVSIM Navtest measured by PDMS. Bold and underlined values denote the best and second-best results in each column.}
\label{tab:pdm-v1-main}
\centering\scriptsize
\setlength{\tabcolsep}{6pt}
\renewcommand{\arraystretch}{1.02}
\begin{tabular*}{\linewidth}{@{\extracolsep{\fill}}lrrrrrc@{}}
\toprule
Method & NC$\uparrow$ & DAC$\uparrow$ & TTC$\uparrow$ & Comf.$\uparrow$ & EP$\uparrow$ & PDMS$\uparrow$ \\
\midrule
\multicolumn{7}{@{}l}{\textit{End-to-end planners}} \\
TransFuser~\citep{Chitta2023PAMI} & 97.7 & 92.8 & 92.8 & \best{100.0} & 79.2 & 84.0 \\
DiffusionDrive~\citep{Liao2025DiffusionDrive} & 98.2 & 96.2 & 94.7 & \best{100.0} & 82.2 & 88.1 \\
WoTE~\citep{Li2025WOTE} & 98.5 & 96.8 & 94.9 & \underline{99.9} & 81.9 & 88.3 \\
DriveDPO~\citep{Shang2025DriveDPO} & 98.5 & 98.1 & 94.8 & \underline{99.9} & 84.3 & 90.0 \\
\midrule
\multicolumn{7}{@{}l}{\textit{Vision-language-action models}} \\
FSDrive~\citep{zeng2025fsdrive} & 98.2 & 93.8 & 93.3 & \underline{99.9} & 80.1 & 85.1 \\
AutoVLA~\citep{zhou2025autovla} & 98.4 & 95.6 & \best{98.0} & \underline{99.9} & 81.9 & 89.1 \\
DriveVLA-W0~\citep{li2025drivevlaw0} & 98.7 & \best{99.1} & 95.3 & 99.3 & 83.3 & 90.2 \\
AdaThinkDrive~\citep{luo2025adathinkdrive} & 98.4 & 97.8 & 95.2 & \best{100.0} & 84.4 & 90.3 \\
Curious-VLA~\citep{chen2026curiousvla} & 98.4 & 96.9 & \underline{97.9} & 98.1 & \best{88.5} & 90.3 \\
DriveTeach-VLA~\citep{yang2026driveteach} & 98.5 & 96.9 & \underline{97.9} & 98.2 & \best{88.5} & 90.4 \\
AutoDrive-P$^3$~\citep{ye2026autodrivep3} & \underline{99.1} & 97.4 & 96.5 & \best{100.0} & \underline{84.8} & \underline{90.6} \\
\midrule
\textbf{FocusDrive (SFT)} & 98.7 & 96.8 & 95.5 & \best{100.0} & 82.4 & 88.9 \\
\textbf{FocusDrive (RL)} & \best{99.3} & \underline{98.8} & 97.7 & \best{100.0} & 82.8 & \best{91.1} \\
\bottomrule
\end{tabular*}
\end{table*}

\begin{table*}[t]
\caption{Planning results on the same NAVSIM Navtest split measured by EPDMS. Bold and underlined values denote the best and second-best results in each column.}
\label{tab:epdms-v2-main}
\centering\small
\setlength{\tabcolsep}{6pt}
\renewcommand{\arraystretch}{1.08}
\resizebox{\linewidth}{!}{%
\begin{tabular}{lrrrrrrrrrc}
\toprule
Method & NC$\uparrow$ & DAC$\uparrow$ & DDC$\uparrow$ & TLC$\uparrow$ & EP$\uparrow$ & TTC$\uparrow$ & LK$\uparrow$ & HC$\uparrow$ & EC$\uparrow$ & EPDMS$\uparrow$ \\
\midrule
\multicolumn{11}{l}{\textit{End-to-end planners}} \\
TransFuser~\citep{Chitta2023PAMI} & 96.9 & 89.9 & 97.8 & 99.7 & 87.1 & 95.4 & 92.7 & \underline{98.3} & 87.2 & 84.0 \\
WoTE~\citep{Li2025WOTE} & 98.5 & 96.8 & 98.8 & \underline{99.8} & 86.1 & 97.9 & 95.5 & \underline{98.3} & 82.9 & 87.7 \\
DiffusionDrive~\citep{Liao2025DiffusionDrive} & 98.2 & 96.2 & \underline{99.5} & \underline{99.8} & 87.4 & 97.3 & 96.9 & \best{98.4} & \best{87.7} & 88.2 \\
\midrule
\multicolumn{11}{l}{\textit{Vision-language-action models}} \\
DriveVLA-W0~\citep{li2025drivevlaw0} & 98.5 & \best{99.1} & 98.0 & 99.7 & 86.4 & 98.1 & 93.2 & 97.9 & 58.9 & 86.1 \\
DriveWorld-VLA~\citep{jia2026driveworldvla} & 98.6 & \best{99.1} & \best{99.6} & \underline{99.8} & 87.4 & 97.9 & 97.0 & 97.8 & 78.6 & 86.8 \\
Latent-WAM~\citep{wang2026latentwam} & 98.1 & 97.3 & \best{99.6} & \underline{99.8} & \underline{87.7} & 97.3 & \underline{97.6} & 98.1 & 87.3 & 89.3 \\
DriveFuture~\citep{hong2026drivefuture} & \underline{99.1} & 97.2 & 99.3 & \underline{99.8} & \best{87.8} & \underline{98.5} & 96.1 & \underline{98.3} & 85.2 & \underline{89.9} \\
\midrule
\textbf{FocusDrive (SFT)} & 98.7 & 96.8 & \underline{99.5} & \best{99.9} & 86.7 & 98.0 & \best{97.8} & \underline{98.3} & \underline{87.5} & 89.1 \\
\textbf{FocusDrive (RL)} & \best{99.3} & \underline{98.8} & \best{99.6} & \best{99.9} & 84.9 & \best{98.9} & 97.1 & \best{98.4} & 87.2 & \best{91.1} \\
\bottomrule
\end{tabular}}
\end{table*}
}}}}}}}

\afterpage{\afterpage{\afterpage{\afterpage{\afterpage{\afterpage{\afterpage{\afterpage{\begin{figure*}[t]
\centering
\includegraphics[width=0.88\linewidth]{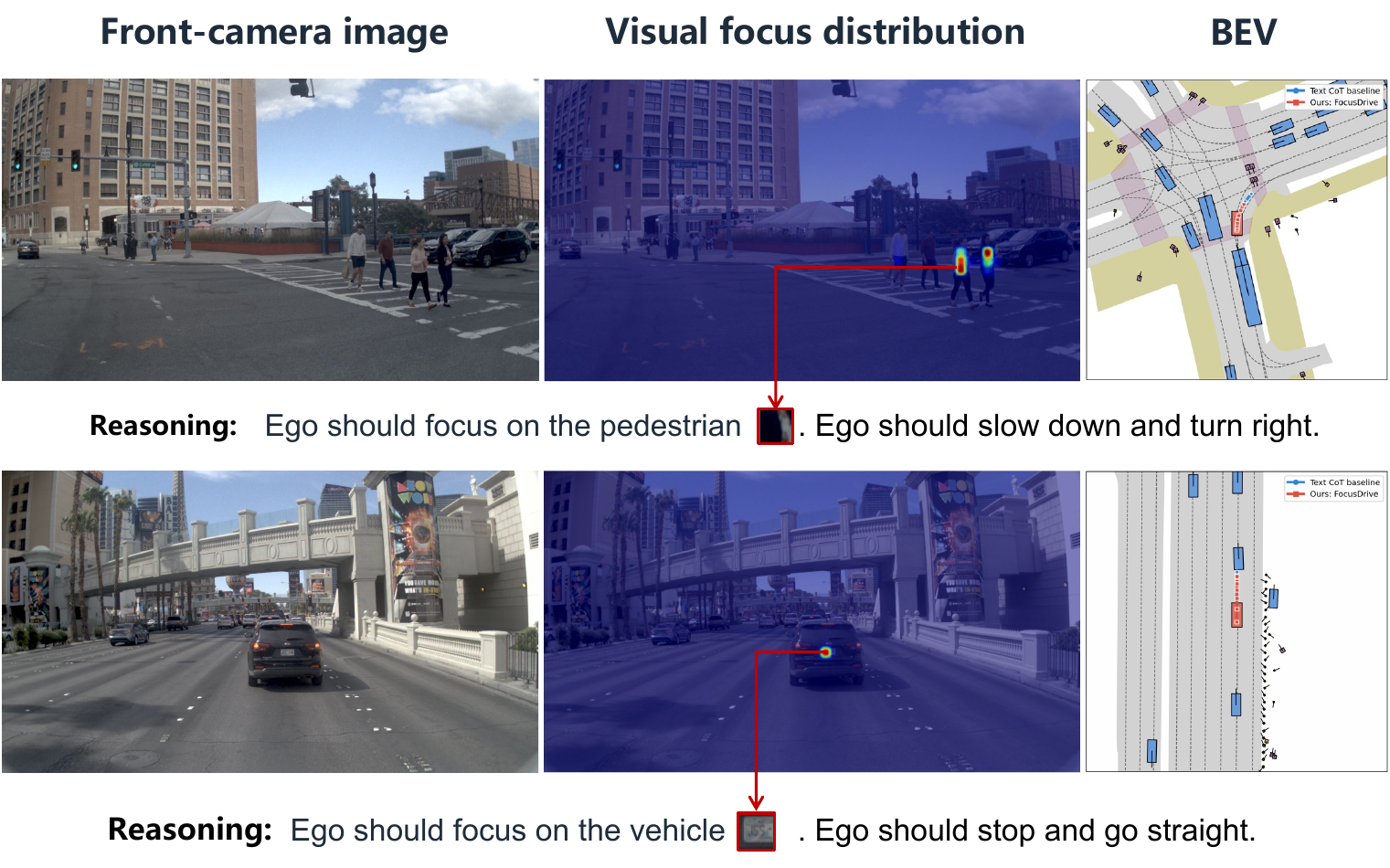}
\caption{\textbf{Visual focus for planning.} Scene images, visual focus distributions, and predicted trajectories with FocusDrive's reasoning. The text CoT baseline (blue) receives an NC score of 0 in the top case and a TTC score of 0 in the bottom case; FocusDrive (red) avoids the corresponding penalties in both cases.}
\label{fig:planning-qualitative}
\end{figure*}

\begin{table}[t]
\begin{minipage}[t]{0.49\linewidth}
\vspace{0pt}
\caption{Comparison with our text CoT baseline under SFT and reinforcement fine-tuning. $\Delta$ denotes the gain over the baseline in the corresponding metric and training stage.}
\label{tab:ablation-representations}
\centering\scriptsize
\setlength{\tabcolsep}{2pt}
\renewcommand{\arraystretch}{1.12}
\begin{tabular*}{\linewidth}{@{\extracolsep{\fill}}lrrrr@{}}
\toprule
Method & PDMS$\uparrow$ & $\Delta$ & EPDMS$\uparrow$ & $\Delta$ \\
\midrule
\multicolumn{5}{@{}l}{\textit{Supervised fine-tuning}} \\
Text CoT baseline & 87.6 & -- & 87.7 & -- \\
\textbf{FocusDrive} & \best{88.9} & \best{+1.3} & \best{89.1} & \best{+1.4} \\
\midrule
\multicolumn{5}{@{}l}{\textit{Reinforcement fine-tuning}} \\
Text CoT baseline & 90.2 & -- & 90.1 & -- \\
\textbf{FocusDrive} & \best{91.1} & \best{+0.9} & \best{91.1} & \best{+1.0} \\
\bottomrule
\end{tabular*}
\end{minipage}\hfill
\begin{minipage}[t]{0.48\linewidth}
\vspace{0pt}
\caption{Focus representations and components under SFT. Ref. input: coordinates (Text), learned ID embeddings (ID), or image-conditioned patch prototypes (Patch ref.).}
\label{tab:ablation-components}
\centering\scriptsize
\setlength{\tabcolsep}{1.7pt}
\renewcommand{\arraystretch}{1.31}
\begin{tabular*}{\linewidth}{@{\extracolsep{\fill}}lccrr@{}}
\toprule
Method & Warm-up & Ref. input & PDMS$\uparrow$ & EPDMS$\uparrow$ \\
\midrule
Text-2D & -- & Text & 88.2 & 88.5 \\
FocusDrive & -- & Patch ref. & 88.6 & 88.9 \\
\midrule
Text-2D & $\checkmark$ & Text & 88.3 & 88.5 \\
FocusDrive & $\checkmark$ & ID & 88.6 & 88.8 \\
\textbf{FocusDrive} & $\checkmark$ & Patch ref. & \best{88.9} & \best{89.1} \\
\bottomrule
\end{tabular*}
\end{minipage}
\end{table}

}}}}}}}}

\begin{abstract}
Driving decisions depend on both where to focus and how to act on what is seen. Effective driving reasoning must establish which objects matter, where they are, and how they inform the intended action. Text-based rationales can describe a driving response while leaving its correspondence to specific visual evidence implicit. Visual focus provides a concrete starting point for this connection by identifying what matters in the scene and where it is. We propose \textbf{FocusDrive}, a structured multimodal reasoning framework that organizes end-to-end planning around explicit visual focus. It pairs descriptions of decision-relevant objects with image-patch references, bringing explicit visual focus into the reasoning that generates driving plans and trajectories. We first assess this focus representation through driver gaze prediction, then investigate its role in planning reasoning using driving-focus annotations within existing NAVSIM training scenes. Experiments on W3DA and NAVSIM demonstrate competitive gaze prediction and end-to-end planning performance, with FocusDrive improving over text-based chain-of-thought. These results support visual focus as an effective link between scene understanding and driving action.
\end{abstract}

\section{Introduction}
\label{sec:introduction}

Knowing where to focus is an essential part of driving. Visual focus concerns the scene elements selected for consideration as a driver decides how to respond. Human gaze offers an observable perspective on this process, motivating a substantial body of research on driver attention prediction. Early deep saliency models learned spatial attention distributions from convolutional features~\citep{cornia2016mlnet}, while driving-specific approaches incorporated temporal context, critical-event sampling, and semantic scene structure to predict attention in routine and challenging situations~\citep{palazzi2017dreyeve,BDDA,fang2019dada}. Predicted attention has also been used to improve learned driving control~\citep{makrigiorgos2019attention}. Together, these studies establish a foundation for modeling driving focus and understanding how drivers prioritize visual information.

More recent work connects these spatial predictions to language, making the objects and reasons behind attention explicit. LLada jointly models where drivers attend, what they attend to, and why~\citep{zhou2025where}; FSDAM couples attention prediction with structured explanations under limited supervision~\citep{hamid2025fsdam}; and DualGaze-VLM studies text-guided object-level gaze prediction~\citep{ke2026dualgaze}. Together, gaze prediction and its language-based explanations connect where drivers look with the objects and reasons behind that selection. This perspective on human attention motivates our approach to driving reasoning: explicitly identify and locate the relevant scene elements, then reason about how to respond to them.

Intermediate reasoning has become an important means of connecting scene understanding with action generation in driving VLMs and VLAs. Language-based approaches organize scene interpretation into descriptions and structured questions~\citep{sima2023drivelm,tian2024drivevlm,hwang2024emma}, and further developments connect CoT reasoning with action generation~\citep{zhou2025autovla,chen2026curiousvla}. In current driving VLA CoT, however, intermediate reasoning is primarily expressed in natural language. Object descriptions convey what matters and why, but their correspondence to specific regions of the input image remains implicit in the generated sequence. Meanwhile, recent VLM research explores richer spatial and multimodal representations to improve visual understanding, ranging from textual coordinate prediction~\citep{song2026simpleseg} to continuous visual tokens~\citep{qin2025covt} and image-region references~\citep{su2025padt}. These developments offer ways to connect language reasoning more directly with the spatial structure and visual information of an image. We bring this perspective into driving through visual focus: connecting the object under consideration to its image representation before reasoning about the driving response.

We introduce \textbf{FocusDrive}, a structured multimodal CoT framework that organizes driving reasoning around explicit visual focus. FocusDrive pairs an object description with a generated patch reference to express decision-relevant visual focus within driving reasoning, using PaDT's dynamic visual-reference interface~\citep{su2025padt}. The description identifies the scene element under consideration, while the reference connects it to an image region and its visual representation. This pairing expresses where the model focuses and provides image-grounded context for the reasoning that follows. We first apply this formulation to language-related driver gaze prediction, where a lightweight adapter converts patch scores into dense attention heatmaps. This task evaluates how well the representation captures the spatial distribution of driver focus. For end-to-end planning, the policy generates a visual-focus span, an action plan, and future waypoints in sequence. Its object description and patch reference supply spatial and visual context as it reasons about how to respond to the selected scene element, connecting visual focus with action generation.

To support learning this form of reasoning, we develop driving-focus annotations within NAVSIM. Automated risk assessment~\citep{wang2015safetyfield,cheng2026ea} and VLM-based screening identify candidate scenes and objects, concentrating focus annotation on situations that call for it, inspired by AdaThinkDrive's selective reasoning perspective~\citep{luo2025adathinkdrive}. The annotator refines these candidates and selects the final visual focus, bringing human judgments of driving relevance into the spatial supervision. These annotations make human-selected driving focus learnable within planning sequences, pairing object descriptions with spatial references before the subsequent reasoning and action targets.

We evaluate FocusDrive on W3DA~\citep{zhou2025where} and NAVSIM~\citep{dauner2024navsim}, examining both its ability to predict driver gaze and its effectiveness in end-to-end planning. Competitive gaze prediction across normal, safety-critical, and accident scenarios demonstrates the framework's ability to model the spatial distribution of driver attention. We further evaluate the use of explicit visual focus within planning CoT, where FocusDrive achieves competitive end-to-end planning performance on NAVSIM. Its improvement over text-based CoT supports the value of explicitly incorporating driving focus into multimodal reasoning. Ablations further examine how focus representation and visual-reference learning contribute to this improvement.

Our contributions are threefold:
\begin{itemize}
    \item We propose FocusDrive, a structured multimodal CoT framework that organizes driving reasoning around explicit visual focus, connecting object identification and localization with response reasoning and end-to-end planning.
    \item We develop driving-focus annotations within NAVSIM, capturing human selections of decision-relevant objects and their image locations to support learning focus within planning reasoning.
    \item We demonstrate competitive performance in driver gaze prediction and end-to-end planning, and show that visual-focus reasoning improves planning over text-based CoT with focus annotations on a small subset of existing NAVSIM training scenes.
\end{itemize}

\section{Related Work}
\label{sec:related-work}

\paragraph{Driver attention and language-guided gaze prediction.}
Learning where drivers look has long provided a route to understanding visual selection in driving. Convolutional saliency models~\citep{cornia2016mlnet} learn dense spatial predictions from visual features. Driving-specific studies extend this line to temporal scene context in DR(eye)VE~\citep{palazzi2017dreyeve}, critical situations in BDD-A~\citep{BDDA}, and accident scenarios in DADA~\citep{fang2019dada}. The usefulness of predicted human attention for learned driving control has also been explored~\citep{makrigiorgos2019attention}. More recent approaches connect spatial attention to language: LLada jointly predicts attention maps, attended objects, and explanations~\citep{zhou2025where}; FSDAM learns spatial prediction and structured explanations with limited semantic supervision~\citep{hamid2025fsdam}; and DualGaze-VLM introduces text-guided object-level gaze modeling alongside global prediction~\citep{ke2026dualgaze}. Together, these works establish a progression from predicting attention distributions to describing their semantic content.

\paragraph{Reasoning in driving vision-language-action models.}
Research on reasoning in driving VLAs addresses both how scene information is represented and how it is used to produce actions. DriveLM~\citep{sima2023drivelm}, DriveVLM~\citep{tian2024drivevlm}, and EMMA~\citep{hwang2024emma} organize scene understanding and planning through language. RDA-Driver~\citep{huang2024rdadriver} and ORION~\citep{fu2025orion} study the connection between reasoning and action generation, while AutoVLA~\citep{zhou2025autovla} and AdaThinkDrive~\citep{luo2025adathinkdrive} adapt the amount of reasoning to the driving situation. A further line of work enriches the information carried by intermediate representations. CoT-VLA~\citep{zhao2025cotvla} uses future visual goals in robotic manipulation, and FSDrive~\citep{zeng2025fsdrive} introduces future visual scenes into driving CoT. DriveVLA-W0~\citep{li2025drivevlaw0} exploits world-model supervision, OneVL~\citep{lu2026onevl} develops latent reasoning for planning, and DynVLA~\citep{shang2026dynvla} reasons through compact dynamics tokens. DriveTeach-VLA~\citep{yang2026driveteach} uses spatial trajectory prompts to guide policy learning. Together, these approaches establish intermediate reasoning as a means of coupling scene interpretation with action generation, through both language and increasingly rich visual representations.

\section{Method}
\label{sec:method}

FocusDrive organizes driving reasoning around explicitly selected visual focus. The model describes an object, references its location in the image, and continues reasoning toward a driving response with the selected visual features in context. This formulation connects what the model considers with how it plans to act. We describe the construction of driving-focus supervision and the visual-reference mechanism, then present its use in gaze prediction and end-to-end planning.

\subsection{Task Formulation}
\label{sec:task-formulation}

\paragraph{Driver gaze prediction.}
Given a driving image $I$ and a task prompt $q$, the goal is to predict a dense gaze map $\hat{G}\in[0,1]^{H_g\times W_g}$ that matches the recorded driver gaze $G$. We study global attention prediction on W3DA~\citep{zhou2025where}. The language and spatial annotations associated with the training examples provide supervision for relating attended objects to image regions, while the gaze maps describe how attention is distributed over the scene.
The prompt contains a scene description and task instruction.

\paragraph{End-to-end planning.}
The NAVSIM planning task~\citep{dauner2024navsim} provides sensor observations together with a navigation command and ego state. We denote the planning input by $x=(\mathcal{O},c,s)$, where $\mathcal{O}$ contains the available sensor observations, $c$ is the navigation command, and $s$ is the ego state. The output is an ego-frame future trajectory $\tau=\{(x_k,y_k,\theta_k)\}_{k=1}^{K}$, where $\theta_k$ denotes heading. Under the NAVSIM evaluation protocol considered here, the trajectory consists of $K=8$ waypoints spanning four seconds at $0.5$-second intervals.

\subsection{Driving-Focus Annotations}
\label{sec:focus-annotation}

Gaze data provides a foundation for learning where drivers attend. To make visual focus useful for planning, we additionally need spatial annotations that relate scene elements to the driving decision. We construct these driving-focus annotations within the existing NAVSIM training scenes. Each annotation pairs an object description $d_o$ with a human-selected image location $u=(u_x,u_y)\in[0,1]^2$, expressing both the content and location of the selected focus.

The annotation process proceeds from coarse screening to human selection. Drawing on established approaches to driving risk assessment~\citep{hayward1972nearmiss,wang2015safetyfield,cheng2026ea}, we screen for potentially critical agent interactions and combine these physical risk cues with VLM assessments of the scene and its objects. Together, they narrow the scenes and candidate objects requiring human consideration. Inspired by the selective reasoning perspective of AdaThinkDrive~\citep{luo2025adathinkdrive}, we concentrate explicit focus annotation on situations in which selecting a particular object helps explain the driving decision. The annotator then assesses the candidates in context, chooses the object to focus on, and manually marks its image location. The automated stage directs annotation effort, while human judgment determines the final spatial target.

We map each selected location to the VLM's spatial grid and pair its patch reference with the object description to form a focus span. Annotated scenes include this span before the driving plan, while all scenes retain plan and waypoint supervision. Annotation details appear in Appendix~\ref{sec:annotation-details}.

\subsection{Representing Visual Focus}
\label{sec:visual-reference}

FocusDrive expresses visual focus through an object description paired with a generated patch reference. Each reference points to a merged cell in the image grid and carries a visual representation contextualized by the vision encoder.

We implement image-patch references in the VLM using the dynamic visual-reference mechanism of PaDT~\citep{su2025padt}. For an input image $I$, the visual backbone produces spatial embeddings $E(I)=[e_1,\ldots,e_N]^\top\in\mathbb{R}^{N\times d}$, where $N$ is the number of spatial reference units and $d$ is the language model's hidden dimension. We refer to these grid cells as image patches.

A visual-reference projector transforms each patch embedding $e_j$ into a prototype $p_j$ in the VLM's hidden space. It first applies layer normalization, then adds a low-rank residual projection implemented by two bias-free linear layers with a low-dimensional bottleneck. The resulting prototypes form $P(I)=[p_1,\ldots,p_N]^\top\in\mathbb{R}^{N\times d}$. Each prototype serves both as an output choice and as the input embedding of its reference token.

Let $\mathcal{V}_{\mathrm{text}}$ be the ordinary text vocabulary, with output embeddings $W_{\mathrm{text}}\in\mathbb{R}^{|\mathcal{V}_{\mathrm{text}}|\times d}$. We extend its output choices with the image-dependent reference set $\mathcal{V}_{\mathrm{vis}}(I)=\{v_1,\ldots,v_N\}$. At decoding step $t$, the hidden state $h_t\in\mathbb{R}^{d}$ scores both sets:
\begin{equation}
    \ell_t=\begin{bmatrix}W_{\mathrm{text}}h_t\\P(I)h_t\end{bmatrix},\qquad
    p_\theta(y_t\mid\mathcal{C},y_{<t})=
    \frac{\exp(\ell_{t,y_t})}{\sum_{u\in\mathcal{V}_{\mathrm{text}}\cup\mathcal{V}_{\mathrm{vis}}(I)}\exp(\ell_{t,u})}.
    \label{eq:reference-decoding}
\end{equation}
The context $\mathcal{C}$ contains the image and task inputs, and $y_{<t}$ is the generated prefix. Text and references are normalized within the same distribution, allowing both to appear in one autoregressive completion. Only references belonging to the current observation are available to that example.

When a reference $v_j$ is generated, its image-dependent prototype $p_j$ becomes the input embedding at that token position for subsequent decoding. Text tokens retain their learned embeddings.

\subsection{From Visual References to Gaze Prediction}
\label{sec:gaze-learning}

Gaze prediction requires a dense heatmap, so we equip the visual-reference representation with a task-specific gaze adapter. At the position $t^\star$ preceding the reference token, the VLM's final-layer hidden state $h_{t^\star}$ is matched against the image-dependent prototypes. The resulting scores retain the spatial correspondence of the patch grid:
\begin{equation}
    z_{t^\star}=P(I)h_{t^\star}\in\mathbb{R}^{N},\qquad
    \hat{G}=f_\phi\!\left(z_{t^\star}\right).
    \label{eq:gaze-readout}
\end{equation}
The adapter $f_\phi$ normalizes these scores across patches, reshapes them onto the spatial grid, and decodes the dense map through five convolutional layers with bilinear upsampling. We train the VLM and adapter jointly: annotated attention text and spatial references supervise the completion through a language-modeling loss, while recorded gaze maps supervise the heatmap output through binary cross-entropy and KL divergence.

At inference, the image, scene description, and task instruction are followed by a fixed prefix ending at $t^\star$. A single forward pass then produces the heatmap through Eq.~\ref{eq:gaze-readout}, without supplying the ground-truth reference or attention explanation. The adapter architecture and training objective are detailed in Appendix~\ref{sec:gaze-config}, and the prompt templates in Appendix~\ref{sec:gaze-prompts}.

\subsection{From Visual Focus to End-to-End Planning}
\label{sec:focus-planning}

FocusDrive expresses visual focus within the sequence that generates the driving plan. When a focus span is generated, its object description and patch reference provide spatial grounding and image-conditioned visual information for the subsequent action plan and future waypoints.

\subsubsection{Supervised Learning of Focus-Based Reasoning}

To teach the planner this progression, we organize an annotated completion $Y$ as
\begin{equation}
    \underbrace{[d_o,\,v_j]}_{\text{visual focus}}
    \ \longrightarrow\ 
    \underbrace{a}_{\text{action plan}}
    \ \longrightarrow\ 
    \underbrace{w}_{\text{waypoint text}}.
    \label{eq:planning-sequence}
\end{equation}
Here, $d_o$ is the description of the selected object and $v_j$ references its image patch. The action plan $a$ expresses the intended driving behavior, and $w$ represents the future trajectory $\tau$ as numerical text and delimiters encoded by the ordinary text tokenizer. Thus, $a$ and $w$ are both text-token sequences, whereas $v_j$ is an image-dependent visual reference. These spans use \texttt{<focus>}, \texttt{<plan>}, and \texttt{<answer>} delimiters, respectively. See Appendix~\ref{sec:planning-formats} for input and completion formats.

The pretrained VLM already supports language generation, whereas image-patch references introduce a new vocabulary for selecting visual focus. We begin with visual-reference token (VRT) warm-up, training the model to jointly describe and reference objects before learning complete planning sequences. Before selective driving-focus screening, we construct broader object-reference annotations across Navtrain, describing and localizing relevant objects in each annotated scene. Each warm-up completion contains multiple object descriptions, including relative position, and their corresponding patch references, without action-plan or waypoint targets.

Both reference warm-up and planning SFT use token-level cross-entropy, with stage-specific contexts and target completions:
\begin{equation}
    \mathcal{L}_{\mathrm{CE}}(\theta;\mathcal{D})
    =-\mathbb{E}_{(\mathcal{C},Y)\sim\mathcal{D}}
    \sum_{t=1}^{|Y|}\log p_\theta(y_t\mid\mathcal{C},y_{<t}).
    \label{eq:planning-loss}
\end{equation}
Here, $\mathcal{C}$ contains the image and task inputs. During warm-up, $\mathcal{D}=\mathcal{D}_{\mathrm{vrt}}$ and the target contains object descriptions and patch references. During planning SFT, $\mathcal{D}=\mathcal{D}_{\mathrm{plan}}$ and the target contains the focus span, when annotated, followed by the action plan and waypoint text. Scenes without focus annotations still supervise the plan and future motion. Appendices~\ref{sec:planning-formats} and~\ref{sec:planning-training} specify the completion formats and stage-specific training configurations.

\subsubsection{Reinforcement Fine-Tuning with GRPO}

Supervised learning establishes how to express visual focus and continue from it to a driving plan. Reinforcement fine-tuning adds feedback on the resulting behavior, allowing the policy to improve beyond matching the annotated completion. A driving scene can admit several plausible plans, whose quality is not fully captured by matching a single annotated sequence. Following outcome-based policy optimization~\citep{zhou2025autovla,shang2026dynvla,li2025recogdrive}, we use group relative policy optimization (GRPO) to compare sampled completions for the same observation and reinforce those with better driving outcomes.

Let $\pi_\theta$ denote the VLM policy used for planning. For each observation $x$, the current policy samples a group of $G$ completions $\{Y_i\}_{i=1}^{G}$. The waypoint text in each completion is parsed into a candidate trajectory $\tau_i$, and the completion receives the PDMS of this trajectory as its reward, $r_i=\operatorname{PDMS}(\tau_i;x)$. This reward evaluates collision avoidance, road compliance, time to collision, comfort, and progress, connecting the generated reasoning sequence to the quality of its resulting driving behavior. Comparing candidates for the same scene gives a relative advantage
\begin{equation}
    A_i=\frac{r_i-\bar{r}}{\sigma_r+\delta},\qquad
    \bar{r}=\frac{1}{G}\sum_{i=1}^{G}r_i,
    \label{eq:grpo-advantage}
\end{equation}
where $\sigma_r$ is the within-group reward standard deviation and $\delta$ is a numerical stabilizer.

The group-relative loss is
\begin{equation}
    \mathcal{L}_{\mathrm{RL}}(\theta)
    =-\frac{1}{G}\sum_{i=1}^{G}\frac{A_i}{|Y_i|}
    \sum_{t=1}^{|Y_i|}\log\pi_\theta(y_{i,t}\mid x,y_{i,<t})
    +\beta\widehat{\mathcal{K}}(\pi_\theta,\pi_{\mathrm{ref}}).
    \label{eq:grpo-objective}
\end{equation}
Here, $\pi_{\mathrm{ref}}$ is the frozen SFT policy and $\beta$ controls KL regularization. The sampled completions and their advantages are treated as fixed during the policy update. The loss covers the generated reasoning and waypoint tokens, while KL regularization constrains deviation from the SFT reference. Appendix~\ref{sec:planning-training} provides the KL estimator, token masking, and optimization settings.

\section{Experiments}
\label{sec:experiments}

Our experiments examine the progression from modeling visual focus to reasoning with it. Gaze prediction evaluates the spatial accuracy of the focus representation, while planning evaluates its value for action generation. Ablations examine explicit focus supervision, visual-reference preparation, and the use of selected visual features during planning.

\subsection{Experimental Setup}
\label{sec:experimental-setup}

\paragraph{Driver gaze benchmark.}
W3DA~\citep{zhou2025where} provides language-enriched driver gaze data spanning Normal (DR(eye)VE~\citep{DReyeVE} and LBW~\citep{LBW}), Safety (BDD-A~\citep{BDDA}), and Accident (DADA-2000~\citep{DADA-2000}) scenarios. We follow the G-W3DA data preparation~\citep{ke2026dualgaze}, using 39,673 training and 21,985 test frames. Gaze maps are evaluated at $H_g\times W_g=256\times256$. We report KL divergence (KLdiv), correlation (CC), similarity (SIM), and AUC-Judd/AUC-Borji (AUC-J/AUC-B), with Normal results averaged over all DR(eye)VE and LBW test frames.

\paragraph{Planning benchmark.}
NAVSIM~\citep{dauner2024navsim} evaluates end-to-end planning through non-reactive simulation on recorded driving scenes. Following prior driving VLAs~\citep{luo2025adathinkdrive,yang2026driveteach}, we train on Navtrain and evaluate on the 12,146 Navtest scenarios using both PDMS from NAVSIM v1 and EPDMS introduced in NAVSIM v2. The two metrics assess complementary aspects of planning safety, compliance, comfort, and progress.

\paragraph{Implementation details.}
FocusDrive uses Qwen3-VL-2B~\citep{bai2025qwen3vl} with a visual-reference projector of bottleneck width $r=64$. We train the gaze model on 2 GPUs and the planner on 16 GPUs with 96\,GB of memory each. For the reported gaze experiments, we train for three epochs with a per-GPU batch size of 12, using learning rates of $10^{-5}$ (VLM) and $10^{-4}$ (adapter).

For planning, the input comprises a $1920\times1080$ front-camera image, navigation command, and ego velocity and acceleration. We add focus annotations to 13.1\% of the existing Navtrain scenes and retain planning supervision on the full Navtrain set. Appendix~\ref{sec:planning-training} provides the stage-specific optimization, rollout, and reward settings.

\subsection{Main Results}
\label{sec:main-results}

\paragraph{Driver gaze prediction.}
Table~\ref{tab:gaze-main} shows strong performance across Normal, Safety, and Accident scenarios, with FocusDrive leading most reported metrics. It achieves the strongest KLdiv, CC, and AUC results across all three scenario groups, together with the highest SIM in Normal and Accident scenes. The largest correlation gain over DualGaze-VLM occurs in safety-critical scenes, supporting the representation's ability to localize driver focus in demanding situations. These results provide a strong spatial modeling foundation for studying visual focus in planning reasoning.

\paragraph{End-to-end planning.}
FocusDrive reaches 91.1 PDMS and 91.1 EPDMS (Tables~\ref{tab:pdm-v1-main} and~\ref{tab:epdms-v2-main}), leading the explicit CoT planners compared here. The comparison includes text-based reasoning, adaptive thinking, exploration-oriented training, and spatial prompting~\citep{luo2025adathinkdrive,chen2026curiousvla,yang2026driveteach}, as well as perception--prediction--planning chains~\citep{ye2026autodrivep3}. These results support organizing driving reasoning around an explicit connection between the relevant object, its image location, and the intended response. FocusDrive turns this connection into a multimodal reasoning sequence that guides end-to-end planning. Strong collision avoidance and road compliance support the PDMS result. FocusDrive leads NC and TTC and ties for the highest TLC in Table~\ref{tab:epdms-v2-main}, aligning with the annotation emphasis on vehicle interactions and traffic-control cues. Together with the gaze results, these findings support the progression from modeling where drivers attend to using visual focus in end-to-end planning.

\subsection{Ablation Studies and Analysis}
\label{sec:ablations}

\paragraph{Structured reasoning through visual focus.}
Table~\ref{tab:ablation-representations} compares FocusDrive with our text CoT baseline, which follows the language-based reasoning format of Curious-VLA before waypoint generation \citep{chen2026curiousvla}. FocusDrive adds focus supervision and structures the completion around an object description and image-patch reference, followed by the driving plan. This complete formulation improves SFT performance by 1.3 PDMS and 1.4 EPDMS. After reinforcement fine-tuning, the gains remain 0.9 and 1.0 points, respectively, supporting focus-based reasoning as a stronger basis for planning at both stages.

\paragraph{Focus representation and reference preparation.}
Table~\ref{tab:ablation-components} compares focus representations under SFT. Text-2D expresses the focal point using ordinary text coordinates, whereas FocusDrive uses patch references. Both receive planning-focus supervision on the same 13.1\% of Navtrain and planning supervision on the full training set. Both improve over the text CoT baseline, supporting the use of spatially grounded VLM reasoning for driving planning. Patch references exceed Text-2D by 0.4 points on both metrics without reference warm-up and by 0.6 points with warm-up. Reference warm-up further improves FocusDrive by 0.3 PDMS and 0.2 EPDMS.

\paragraph{Image-conditioned patch references.}
To examine the reference input, we replace the selected patch prototype with a learned reference-ID embedding while retaining dynamic patch selection. Both variants retain spatial indexing. A learned ID embedding encodes the grid location with a vector shared across images, whereas a patch prototype combines that spatial correspondence with image-conditioned visual semantics. Given the spatial information already available through the reference ID, using the corresponding patch prototype additionally supplies a visual representation of the selected region in the current image. This image-conditioned input improves PDMS and EPDMS by a further 0.3 points each, indicating a benefit beyond the location information retained by the ID variant. Patch references thus serve both to express the selected focus and to carry its visual representation into subsequent reasoning. Combined with reference warm-up, this formulation achieves the highest scores in Table~\ref{tab:ablation-components}. Together with the larger gains over the text CoT baseline, these comparisons support the complete focus-based formulation: explicit spatial focus improves planning, and the configuration that also incorporates image-conditioned reference inputs performs best among the variants evaluated under SFT.

\paragraph{Qualitative analysis.}
\label{sec:qualitative}

Figure~\ref{fig:planning-qualitative} examines how selected objects, generated reasoning, and driving responses correspond within individual scenes. FocusDrive's generated focus and plan are shown alongside the predicted trajectories of the text CoT baseline and FocusDrive, while patch references identify the image regions explicitly selected by FocusDrive. These examples illustrate how spatial focus is expressed within reasoning and how the resulting plans differ.
Appendices~\ref{sec:gaze-visualization} and~\ref{sec:planning-visualization-appendix} provide gaze comparisons and further planning visualizations, respectively.

\par
\begin{samepage}
\section{Conclusion}
\label{sec:conclusion}

We presented FocusDrive, a framework that makes visual focus an explicit part of driving reasoning. By pairing object descriptions with generated patch references, it connects decision-relevant objects to their visual representations within the planning sequence. Strong gaze-prediction results provide a foundation for modeling focus, while NAVSIM experiments demonstrate the benefits of incorporating explicit focus into driving reasoning. FocusDrive thus connects the study of where to focus with the practical question of how to act on what is seen.
\label{sec:conclusion-end}
\par
\end{samepage}

\subsection*{AI use statement}
We used generative AI tools to assist with code development, literature and reference checking, manuscript drafting and revision, and critical discussion of the methodology and interpretation of experimental results. We also used Qwen3.7-Plus to propose candidate objects during offline driving-focus annotation, as detailed in Appendix~\ref{sec:annotation-details}. These proposals were reviewed by a human annotator, who selected image locations, revised object descriptions, and removed incorrect annotations. AI-assisted manuscript revisions were reviewed and revised by the authors. The authors take full responsibility for the methods, results, claims, and final content of this paper.

\subsection*{Reproducibility statement}
Sections~\ref{sec:method} and~\ref{sec:experimental-setup} describe the model, annotation procedure, training setup, and evaluation metrics. Appendix~\ref{sec:appendix} provides supplementary implementation details to support reproduction of the experiments.

\bibliography{iclr2027_conference}
\bibliographystyle{iclr2027_conference}

\appendix
\section{Appendix}
\label{sec:appendix}

\subsection{Visual Reference Implementation}
\label{sec:reference-implementation}

\paragraph{Reference prototypes.}
The visual backbone supplies spatial embeddings in the language model's hidden space. The projector in Sec.~\ref{sec:visual-reference} uses layer normalization followed by a low-rank residual map:
\begin{equation}
    \bar e_j=\operatorname{LN}(e_j),\qquad
    p_j=\bar e_j+W_{\mathrm{up}}W_{\mathrm{down}}\bar e_j,
    \label{eq:reference-projector}
\end{equation}
where $W_{\mathrm{down}}\in\mathbb{R}^{r\times d}$ and $W_{\mathrm{up}}\in\mathbb{R}^{d\times r}$ are bias-free linear layers, with $r=64$. The current image's prototypes supply both the visual-reference output choices and the input embeddings of the generated reference tokens. Ordinary language tokens retain the VLM's text embeddings.

\paragraph{Mapping locations to patches.}
Each annotated location is assigned to the cell that contains it on the processed image's patch grid. For cell or box annotations, we use the normalized center as the location. Cells are ordered from left to right and top to bottom, so each reference identifies the corresponding visual embedding. The grid is $34\times60$ for planning and $18\times32$ for gaze prediction.

\subsection{Gaze Prediction}
\label{sec:gaze-appendix}

\subsubsection{Model and Training Configuration}
\label{sec:gaze-config}

\paragraph{Inputs.}
Each training and test frame is resized to $1024\times576$ (bilinear), which yields a spatial reference grid of $H_p\times W_p=18\times32$, i.e., $N=576$ reference units. The scenario description is derived from the W3DA question field by truncating it before its question--answer template, so that only the scene description remains and no question or answer text is exposed. Together with the task instruction of App.~\ref{sec:gaze-prompts}, this forms the task prompt $q$.

\paragraph{Training.}
Training pairs each frame with the stage-2 language and bounding-box annotations of the G-W3DA preparation. The completion follows the gaze format of App.~\ref{sec:gaze-prompts}: a fixed focus template containing one reference $v_j$, followed by the annotated attention explanation. The reference index is sampled uniformly from the grid cells whose centers fall inside the annotated box. Training minimizes the language-modeling loss on the completion together with a heatmap loss on the decoded map,
\begin{equation}
    \mathcal{L}=\mathcal{L}_{\mathrm{LM}}+\lambda_g\,\mathcal{L}_{\mathrm{gaze}},\qquad
    \mathcal{L}_{\mathrm{gaze}}=\mathrm{BCE}\bigl(\hat{G},G\bigr)+\lambda_{\mathrm{KL}}\,\mathrm{KL}\bigl(G\,\|\,\hat{G}\bigr),
    \label{eq:gaze-loss}
\end{equation}
where $\mathrm{BCE}$ denotes binary cross-entropy and both maps are normalized to sum to one for the KL term. We set $\lambda_g=3$ and $\lambda_{\mathrm{KL}}=0.5$. We train the gaze model used in evaluation for three epochs with AdamW on 2 GPUs.

\begin{figure}[t]
\centering
\includegraphics[width=\linewidth]{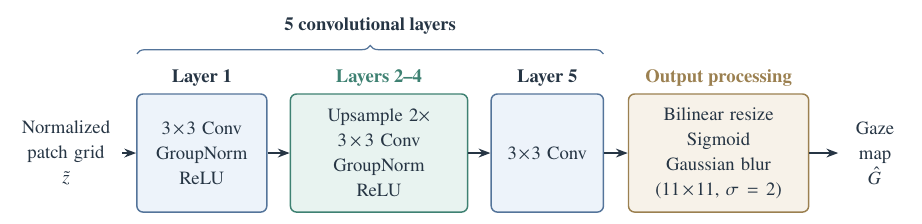}
\caption{Gaze readout architecture. The middle block is repeated three times, giving five convolutional layers in total. Operations within each block run from top to bottom; upsampling uses bilinear interpolation.}
\label{fig:gaze-readout}
\end{figure}

\paragraph{Gaze adapter.}
We obtain $h_{t^\star}$ from the VLM's final layer at the last token of the fixed prefix in App.~\ref{sec:gaze-prompts}, which ends with \texttt{object} followed by a space, immediately before the visual-reference token. Multiplying this state by the prototype matrix gives the reference scores $z_{t^\star}=P(I)h_{t^\star}$. The adapter applies z-score normalization across patches:
\begin{equation}
    \tilde{z}_j=\frac{z_{t^\star,j}-\mu_z}{s_z+10^{-6}},\qquad
    \mu_z=\frac{1}{N}\sum_{j=1}^{N}z_{t^\star,j},\qquad
    s_z=\sqrt{\frac{1}{N-1}\sum_{j=1}^{N}(z_{t^\star,j}-\mu_z)^2}.
    \label{eq:gaze-score-normalization}
\end{equation}
The normalized scores are reshaped onto the patch grid and decoded into a gaze map by the five-layer convolutional readout in Fig.~\ref{fig:gaze-readout}.

\paragraph{Evaluation and inference.}
We report KL divergence (KLdiv), Pearson correlation (CC), similarity (SIM), and AUC-Judd/AUC-Borji (AUC-J/AUC-B). Ground-truth gaze maps are converted to grayscale, resized to $256\times256$, and scaled to $[0,1]$. Let $i$ index pixels, $p_i=\hat{G}_i/\sum_j\hat{G}_j$, and $q_i=G_i/\sum_jG_j$. KLdiv measures distributional discrepancy, while CC measures linear correlation:
\begin{equation}
    \mathrm{KLdiv}=\sum_i q_i\log\frac{q_i}{p_i},\qquad
    \mathrm{CC}=\frac{\sum_i(\hat{G}_i-\overline{\hat{G}})(G_i-\bar{G})}
    {\sqrt{\sum_i(\hat{G}_i-\overline{\hat{G}})^2\sum_i(G_i-\bar{G})^2}},
    \label{eq:gaze-kl-cc}
\end{equation}
where bars denote pixel means. SIM measures the intersection of two unit-sum distributions~\citep{ke2026dualgaze}. Let $M(X)=(X-\min X)/(\max X-\min X)$ denote min--max normalization. With this preprocessing, the normalized-distribution definition is
\begin{equation}
    \mathrm{SIM}=\sum_i\min\!\left(
    \frac{M(\hat{G})_i}{\sum_j M(\hat{G})_j},\,
    \frac{M(G)_i}{\sum_j M(G)_j}\right).
    \label{eq:gaze-sim}
\end{equation}
Numerical stabilizers are omitted from the metric formulas for clarity.

For AUC, fixation pixels satisfy $M(G)_i\geq0.7$. Thresholding $M(\hat{G})$ gives the true-positive rate $\mathrm{TPR}$ and false-positive rate $\mathrm{FPR}$. AUC-J uses all non-fixation pixels as negatives, whereas AUC-B uses 100 sets of uniformly sampled image pixels as negatives, each matching the number of fixation pixels:
\begin{equation}
    \mathrm{AUC\text{-}J}=\int\mathrm{TPR}\,d\mathrm{FPR}_{\mathrm{all}},\qquad
    \mathrm{AUC\text{-}B}=\frac{1}{100}\sum_{b=1}^{100}\int\mathrm{TPR}\,d\mathrm{FPR}_{b}.
    \label{eq:gaze-auc}
\end{equation}
Both areas are computed by trapezoidal integration; AUC-J uses prediction values at fixation pixels as thresholds, and AUC-B uses thresholds from $0.1$ to $0.9$ in steps of $0.1$. Metrics are averaged over test frames, with Normal pooling DR(eye)VE and LBW. Given the image and the prompt and prefix in App.~\ref{sec:gaze-prompts}, a single forward pass produces the predicted heatmap through Eq.~\ref{eq:gaze-readout}.

\subsubsection{Prompt and Output Formats}
\label{sec:gaze-prompts}

The gaze task uses a single user message containing the image and the instruction below, where \texttt{<image>} marks the image placeholder and \texttt{\{scene\}} the scenario description of App.~\ref{sec:gaze-config}. This user message is identical in training and evaluation; only the assistant-side continuation differs.

\begin{center}
\fbox{\begin{minipage}{0.96\linewidth}\small\ttfamily\raggedright
<image> is the front view. \{scene\} You are the driver. Locate your primary attention focus: when a safety-critical object needs attention, output <focus>Ego should focus on the object <|VRT\_n|>.</focus> with exactly one VRT reference, then output the attention explanation inside <plan></plan>.
\end{minipage}}
\end{center}

Training completions place the focus span and the annotated explanation inside a reasoning block delimited by \texttt{<think>}; \texttt{<|VRT\_j|>} is rewritten at runtime to the reference index sampled from the annotated box (App.~\ref{sec:gaze-config}), and \texttt{\{explanation\}} is the annotated attention explanation.

\begin{center}
\fbox{\begin{minipage}{0.96\linewidth}\small\ttfamily\raggedright
<think><focus>Ego should focus on the safety-critical object <|VRT\_j|>.</focus><plan>\{explanation\}</plan></think>
\end{minipage}}
\end{center}

At test time, after the chat template and generation prompt, the following fixed scaffold is appended. It selects the same pre-reference decoding position supervised during training but contains neither a reference index nor any other spatial information. The scaffold ends with a space:

\begin{center}
\fbox{\begin{minipage}{0.96\linewidth}\small\ttfamily\raggedright
<think><focus>Ego should focus on the safety-critical object
\end{minipage}}
\end{center}

\subsubsection{Qualitative Results}
\label{sec:gaze-visualization}

\begin{figure*}[t]
\centering
\includegraphics[width=\textwidth]{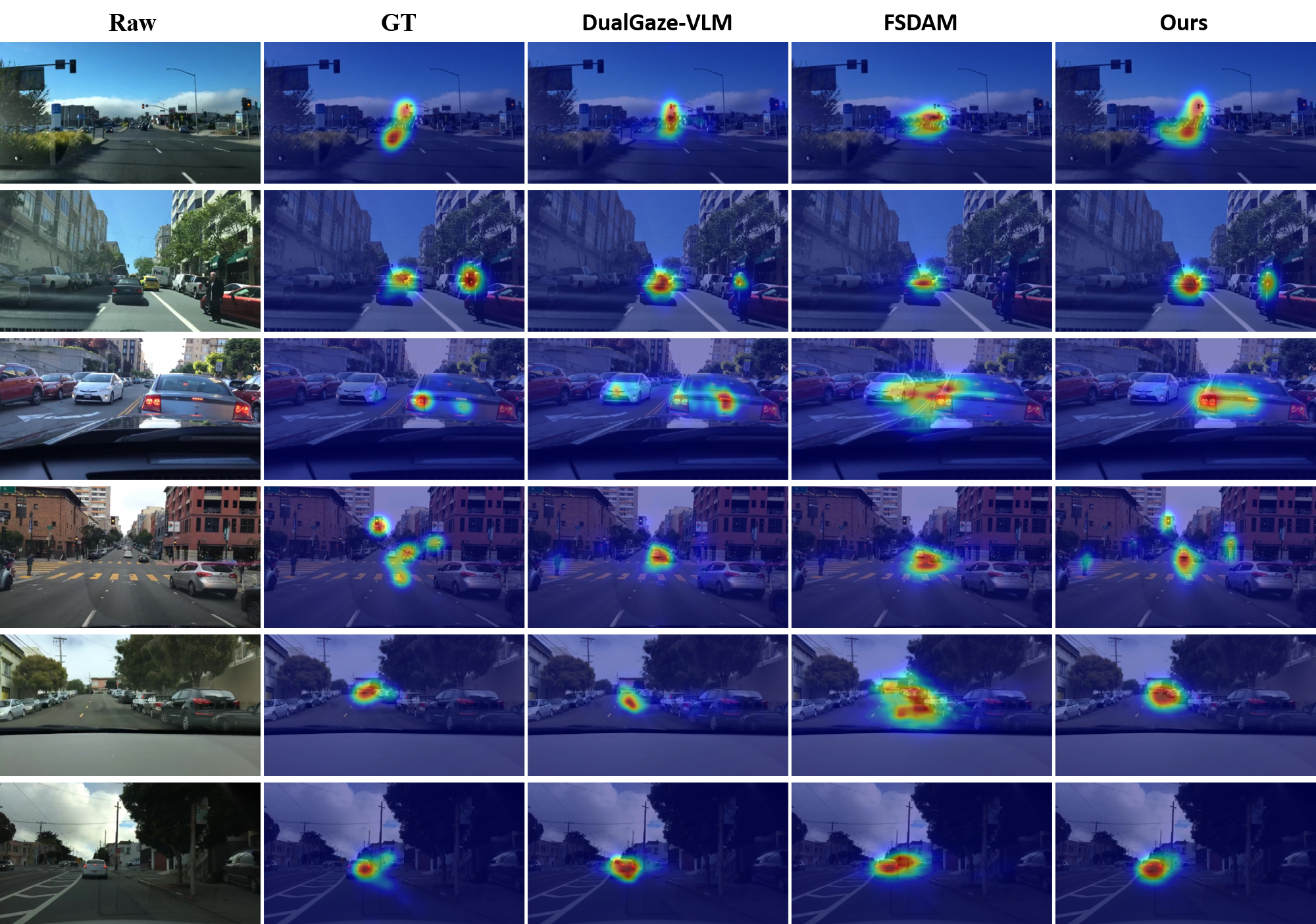}
\caption{Qualitative gaze comparisons on six safety-critical BDD-A test scenes. Columns from left to right: input scene, ground-truth gaze, DualGaze-VLM, FSDAM, and FocusDrive (ours). FocusDrive predictions are produced by the readout of Eq.~\ref{eq:gaze-readout} without ground-truth focus references or attention explanations.}
\label{fig:gaze-qualitative}
\end{figure*}

Figure~\ref{fig:gaze-qualitative} provides example-level evidence complementary to the aggregate results in Table~\ref{tab:gaze-main}. Across the selected scenes, FocusDrive places its dominant responses on the same traffic participants as the ground truth and recovers the separated multi-region structure visible in the first, second, and fourth rows. FSDAM often produces broader or merged responses, as in the first, third, and fifth rows, while DualGaze-VLM weakens or misses secondary regions in several multi-object scenes. The final two rows further show that FocusDrive remains spatially compact in single-target cases. These examples do not replace quantitative evaluation, but illustrate spatial selectivity and the preservation of multiple relevant attention regions without supplying ground-truth focus references or attention explanations.

\subsection{End-to-End Planning}
\label{sec:planning-appendix}

\subsubsection{Driving-Focus Annotations}
\label{sec:annotation-details}

\paragraph{Candidate construction.}
We construct focus annotations within Navtrain through risk-based screening and VLM-assisted candidate selection. Screening combines cues from time to collision, safety-field formulations, and interaction risk assessment~\citep{hayward1972nearmiss,wang2015safetyfield,cheng2026ea}. Qwen3.7-Plus complements these cues in traffic-control and yielding situations, using the front image, candidate boxes, and recorded future ego path to propose relevant objects. The recorded future ego path is used only for offline annotation and is not included in the planning policy's observation input during training or inference.

\paragraph{Human selection and consolidation.}
A single annotator reviews the proposed objects, manually selects their image locations, revises object categories and descriptions, and removes incorrect annotations. The interface supports patch selection on a $34\times60$ image grid. Future work will extend annotation to multiple annotators to assess agreement in focus selection.

Focus annotations cover 13.1\% of Navtrain; the remaining 86.9\% of examples supervise the plan and trajectory without a focus span. Among the focus labels, 37.0\% come from reviewed risk-based candidates and 63.0\% from reviewed traffic-control/yielding candidates. All training examples remain available for planning. We annotate only Navtrain and provide no focus annotations for Navtest, which is reserved for evaluation.

\begin{table}[t]
\caption{Object-category distribution among training focus annotations.}
\label{tab:focus-annotation-distribution}
\centering\small
\setlength{\tabcolsep}{10pt}
\begin{tabular}{lr}
\toprule
Object category & Share (\%) \\
\midrule
Vehicle & 84.43 \\
Traffic light & 7.18 \\
Stop sign & 5.32 \\
Pedestrian & 2.92 \\
Bicycle & 0.10 \\
Other & 0.04 \\
\bottomrule
\end{tabular}
\end{table}

Table~\ref{tab:focus-annotation-distribution} shows that vehicles account for 84.43\% of the selected focus objects, followed by traffic lights and stop signs. The associated plans involve deceleration in 89.2\% of focus-annotated examples. These annotations emphasize collision avoidance and traffic-rule compliance, including potential conflicts and control elements that warrant explicit consideration. They also include scenes flagged for potential risk in which the recorded ego trajectory does not decelerate, so focus selection is not restricted to observed braking. Planning supervision remains available on the full Navtrain set.

\subsubsection{Prompt and Output Formats}
\label{sec:planning-formats}

The user message contains the front-camera image, navigation command, ego velocity and acceleration, with no system turn. The complete planning prompt is reproduced below; braces denote fields filled at runtime.

\begin{center}
\fbox{\begin{minipage}{0.96\linewidth}\small\ttfamily\raggedright
<image> is the front view. Command: \{COMMAND\}. \{ego\_state\_text\}\\
Before the trajectory, output a required <plan>Ego should ...</plan> inside <think></think>. When a safety-critical object needs attention, prepend <focus>Ego should focus on the object <|VRT\_n|>.</focus> with exactly one VRT reference. Do not emit a focus slot otherwise. Output the predicted trajectory in <answer></answer>.\\
For the content in <answer></answer>, Only generate the predicted future waypoints in pure text format:\\{}
[x\_1, y\_1, heading\_1], [x\_2, y\_2, heading\_2], ..., [x\_8, y\_8, heading\_8].\\
Output exactly 8 waypoints in the format [x, y, heading], with numbers in decimal form and exactly 2 digits after the decimal point. Separate waypoints using comma. Do not include extra text, brackets, or invalid values. The output must strictly follow this structure.
\end{minipage}}
\end{center}

The assistant places reasoning inside \texttt{<think>} and waypoints inside \texttt{<answer>}. The schematic completion below illustrates a focus-annotated example; the ellipsis abbreviates the waypoint sequence. Examples without focus annotations omit the focus span and retain the plan and waypoints.

\begin{center}
\fbox{\begin{minipage}{0.96\linewidth}\small\ttfamily\raggedright
<think><focus>Ego should focus on the red traffic light <|VRT\_j|>.</focus><plan>Ego should slow down and go straight.</plan></think>\\
<answer>[x\_1, y\_1, heading\_1], ..., [x\_8, y\_8, heading\_8]</answer>
\end{minipage}}
\end{center}

The plan expresses speed and directional meta-actions obtained from the open-source release of Curious-VLA~\citep{chen2026curiousvla}. Text-2D uses the same reviewed focus categories and action plans, expressing the selected location as ordinary text coordinates. The text CoT baseline uses language reasoning followed by the same waypoint targets. In planning SFT, cross-entropy supervises the complete assistant sequence, including visual references; user-prompt tokens are masked. Waypoint supervision uses text-token cross-entropy.

At inference, the model generates whether to include a focus span as part of its completion. Motivated by selective visual focus, we use an optional single reference to highlight a primary decision-relevant object. The reported NAVSIM gains are achieved with this single-focus format, while the model retains access to the full scene. Future work will examine whether multiple explicit references help in more complex interactions.

\paragraph{Learned reference-ID control.}
The control in Table~\ref{tab:ablation-components} retains image-dependent output selection: reference logits are computed from the current image's patch prototypes. The selected token's input embedding is instead obtained from a learned table with one vector for each of the 2,040 reference IDs, shared across images. This preserves the reference's spatial index while replacing its image-conditioned input embedding. The model retains the full image input.

\subsubsection{Training Configuration}
\label{sec:planning-training}

\paragraph{Reference warm-up and supervised fine-tuning.}
Before selective driving-focus screening, we construct broader object-reference annotations across Navtrain, describing and localizing relevant objects in each annotated scene. Its annotations pair an object description, including relative position, with the corresponding image patch; objects need not be those selected as driving focus for planning. Each warm-up completion jointly generates multiple object descriptions and their corresponding patch references, learning to associate object language with spatial visual content. Two epochs of this warm-up precede three epochs of planning SFT. Planning SFT uses full-parameter AdamW with learning rate $4\times10^{-5}$, weight decay $0.05$, and an effective batch size of 64 (16 GPUs, two examples per GPU, and two gradient-accumulation steps). The schedule uses 100 learning-rate warm-up steps followed by cosine decay.

The warmed Text-2D control starts from the PaDT reference warm-up and then expresses the selected location as normalized text coordinates. The direct Text-2D control starts from the pretrained VLM. The learned reference-ID variant uses its corresponding object-reference warm-up, retaining learned input embeddings for reference tokens throughout training. All planning variants retain the full Navtrain set and three planning-SFT epochs.

\paragraph{Group-relative reinforcement fine-tuning.}
We initialize the policy from planning SFT and keep a frozen SFT reference for KL regularization. Each scene receives $G=16$ sampled completions, one per GPU, with temperature $1.1$, top-$p=1$, and at most 512 new tokens. Rewards are unscaled PDM scores; advantages use a stabilizer of $10^{-4}$, and scene groups with reward standard deviation below $0.05$ are excluded from the update. AdamW uses learning rate $10^{-5}$, no weight decay, gradient-norm clipping at 1.0, and 4,000 optimization steps. The KL coefficient is $0.04$.

The policy and KL losses cover all generated completion tokens, including focus references, language plans, and waypoint text, while excluding padding and positions after EOS. For a sampled token, the KL estimator is $\exp(\Delta_{i,t})-\Delta_{i,t}-1$, where $\Delta_{i,t}=\log\pi_{\mathrm{ref}}(y_{i,t}\mid x,y_{i,<t})-\log\pi_\theta(y_{i,t}\mid x,y_{i,<t})$. Trajectory parsing pads incomplete tails when at least one waypoint is available, and reward evaluation uses the resulting trajectory. Metric-cache or scoring failures receive zero reward.

We use the on-policy log-probability loss defined in Eq.~\ref{eq:grpo-objective}, without policy-ratio clipping. Advantages are fixed during each update, and both the policy and KL terms use the completion-token mask described above.

\paragraph{Evaluation protocol and implementation.}
At evaluation, we use deterministic decoding to produce one trajectory per scene, without stochastic sampling, best-of-$N$ selection, or reranking. Our EPDMS table uses results from the updated NAVSIM implementation with human-penalty filtering enabled, including the official correction to score aggregation after filtering.\footnote{\url{https://github.com/autonomousvision/navsim/issues/151}} Some papers report earlier-version results that may not align with the updated values used in our table. We follow the updated evaluation version consistently.

\clearpage
\subsubsection{Qualitative Results}
\label{sec:planning-visualization-appendix}
\begin{figure}[!ht]
\centering
\includegraphics[width=\textwidth,height=0.82\textheight,keepaspectratio]{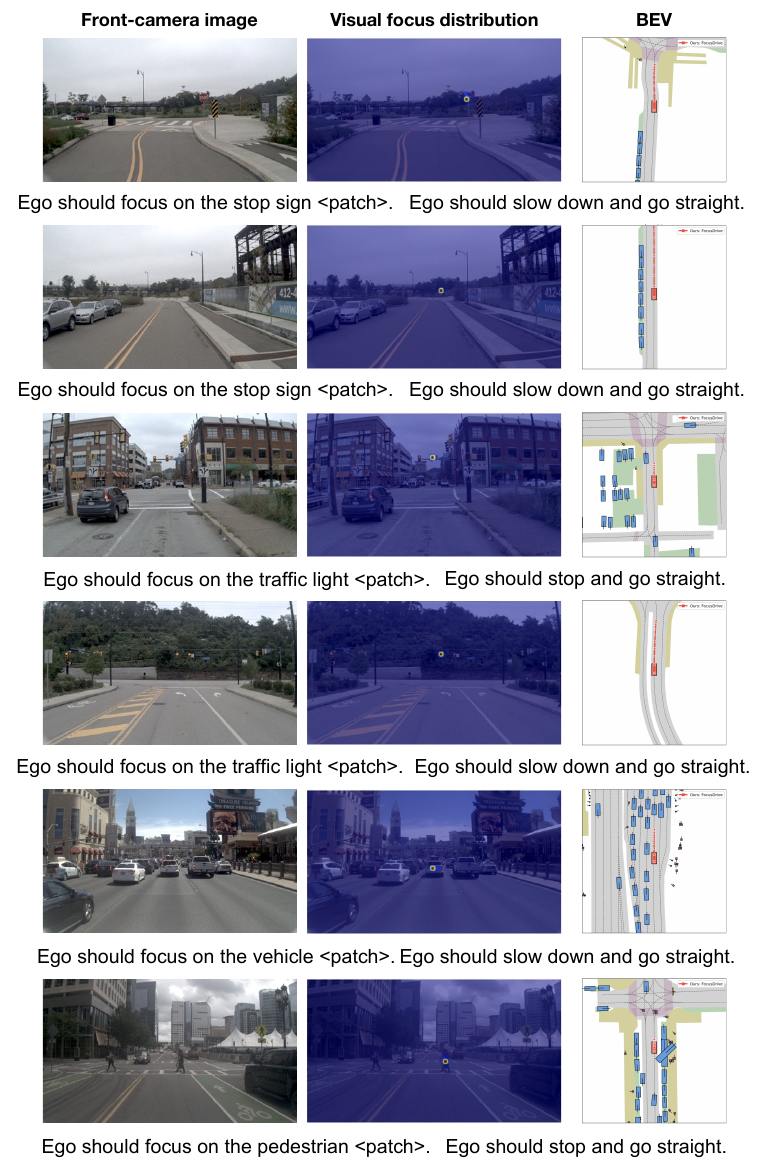}
\caption{Additional planning examples involving stop signs, traffic lights, a vehicle, and a pedestrian. Each row pairs the front-camera image with its visual focus distribution, BEV trajectory, and generated focus and action plan. Red squares denote FocusDrive trajectories; \texttt{<patch>} denotes a selected image-patch reference. Heatmaps visualize the softmax distribution over image-patch reference logits at the reference decoding step, displayed on the patch grid with per-scene color scales.}
\label{fig:planning-qualitative-appendix}
\end{figure}

\end{document}